\documentclass[conference]{IEEEtran}
\IEEEoverridecommandlockouts
\usepackage{cite}
\usepackage{amsmath,amssymb,amsfonts}
\usepackage{algorithmic}
\usepackage{graphicx}
\usepackage{textcomp}
\usepackage{xcolor}
\usepackage[numbers,sort&compress]{natbib}
\usepackage{booktabs}
\usepackage{fontawesome}
\usepackage{tabularx}
\usepackage{multirow}
\usepackage{makecell}
\usepackage{amsmath}
\usepackage{hyperref}
\usepackage{xr}

\begin{document}

\title{From Handcrafted Features to LLMs: A Brief Survey for Machine Translation Quality Estimation}

% \and
% \and
% \and
% \and
% \and

\author{\IEEEauthorblockN{Haofei Zhao\IEEEauthorrefmark{4}\thanks{\IEEEauthorrefmark{1}~Work done during an internship at Huawei.}\IEEEauthorrefmark{1}, Yilun Liu\textsuperscript{\faEnvelope}\thanks{\textsuperscript{\faEnvelope} Yilun Liu is the corresponding author (liuyilun3@huawei.com).}\IEEEauthorrefmark{2}, Shimin Tao\IEEEauthorrefmark{2}, Weibin Meng\IEEEauthorrefmark{2}, Yimeng Chen\IEEEauthorrefmark{2}\\Xiang Geng\IEEEauthorrefmark{3}, Chang Su\IEEEauthorrefmark{2}, Min Zhang\IEEEauthorrefmark{2}, Hao Yang\IEEEauthorrefmark{2}}
\IEEEauthorblockA{\IEEEauthorrefmark{4}School of Computer Science and Engineering, Northeastern University, Shenyang, China}
\IEEEauthorblockA{\IEEEauthorrefmark{2}Huawei, China}
\IEEEauthorblockA{\IEEEauthorrefmark{3}National Key Laboratory for Novel Software Technology, Nanjing University, Nanjing, China}
\IEEEauthorblockA{2272116@stu.neu.edu.cn, \{liuyilun3, taoshimin, mengweibin3, chenyimeng,\\suchang8, zhangmin186, yanghao30\}@huawei.com, gx@smail.nju.edu.cn}}

\maketitle

\begin{abstract}
Machine Translation Quality Estimation (MTQE) is the task of estimating the quality of machine-translated text in real time without the need for reference translations, which is of great importance for the development of MT. After two decades of evolution, QE has yielded a wealth of results. This article provides a comprehensive overview of QE datasets, annotation methods, shared tasks, methodologies, challenges, and future research directions. It begins with an introduction to the background and significance of QE, followed by an explanation of the concepts and evaluation metrics for word-level QE, sentence-level QE, document-level QE, and explainable QE. The paper categorizes the methods developed throughout the history of QE into those based on handcrafted features, deep learning, and Large Language Models (LLMs), with a further division of deep learning-based methods into classic deep learning and those incorporating pre-trained language models (LMs). Additionally, the article details the advantages and limitations of each method and offers a straightforward comparison of different approaches. Finally, the paper discusses the current challenges in QE research and provides an outlook on future research directions.
% This and the IEEEtran.cls file define the components of your paper [title, text, heads, etc.]. *CRITICAL: Do Not Use Symbols, Special Characters, Footnotes, 
% or Math in Paper Title or Abstract.
\end{abstract}


\section{Introduction}

As a critical subfield within NLP, MT has witnessed groundbreaking developments with the advent of deep learning technologies. Nonetheless, the quality of MT remains inherently uncertain. Traditional evaluation metrics, such as BLEU \cite{papineni2002bleu}, METEOR \cite{banerjee2005meteor}, TER \cite{snover2006study}, and CHRF rely on reference translations to assess translation quality. In contrast, QE techniques are capable of automatically evaluating the quality of translations without the need for reference, offering a valuable alternative for appraising the performance of MT systems.

In real-world application scenarios, the use of MT systems often operates without the availability of reference translations. In such contexts, the importance of QE is particularly highlighted. Without access to references, QE provides a crucial independent assessment of translation quality for users, developers, and translation service providers alike. For users, this allows them to more accurately determine the level of translation quality; for developers, QE serves as a robust means of measuring MT system performance; and for translation service providers, QE offers a way to filter out low-quality translations before delivery. These applications demonstrate the extensive applicability and critical role of QE across various levels and sectors within the field.

In the initial stages of QE for MT, there was no unified and clear definition of the field, and the research primarily concentrated on statistical machine translation systems. In 2009, researchers such as Specia et al. \cite{specia2009estimating} introduced an innovative QE framework, which involved manual scoring annotations on translation, the implementation of feature engineering, and the use of ML algorithms to train models capable of predicting translation quality. Since the Workshop on Machine Translation (WMT) established QE as a separate task in 2012, the research has evolved into three main methods: the first is handcrafted feature-based QE; the second leverages deep learning for QE, which further includes subdivisions such as classic deep learning approaches and those incorporating pre-trained LMs; and the third, an emerging method, is based on LLMs. The development of these methods has significantly advanced the progress of QE and gradually improved the accuracy of QE models' assessments. We have categorized these methods just to facilitate the description of the development trend of QE, and cannot guarantee there is no overlap between the methods in different categories.

\begin{figure*}[htbp]
    \centering
    \includegraphics[width=\textwidth]{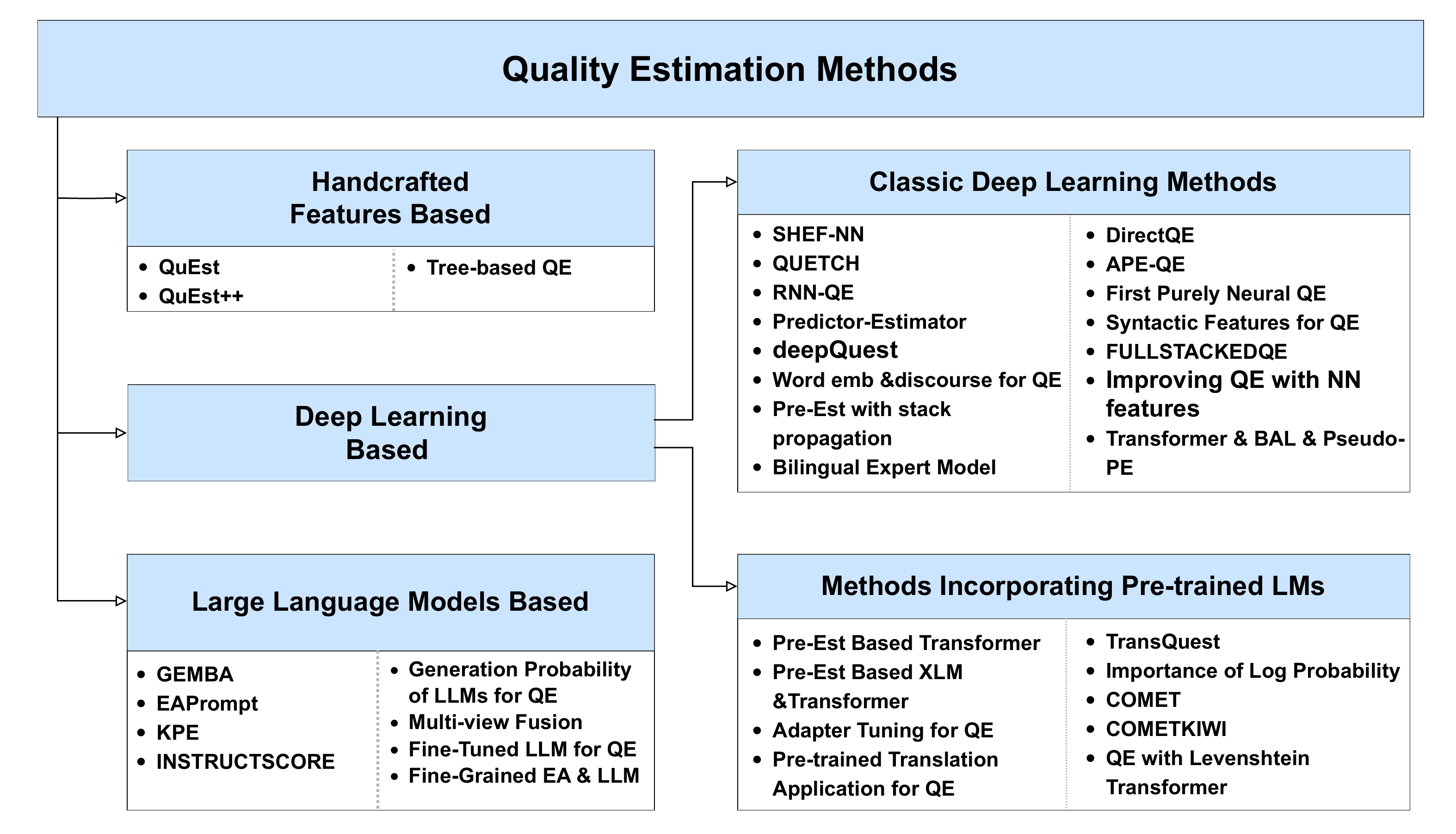}
    \caption{All quality estimation methods investigated in this paper, categorized into handcrafted features-based methods, deep-learning-based methods, and LLM-based methods. Deep-learning-based methods are further divided into classic deep learning methods and those incorporating pre-trained LMs.}
    \label{fig:all}
\end{figure*}

Undoubtedly, approaches based on LLMs have become a focal point of research within the QE domain. Researchers are seeking to harness the extensive knowledge base and learning capabilities of LLMs to achieve new breakthroughs in QE studies. Currently, QE research based on LLMs mainly encompasses the following directions: first, direct prediction of translation quality scores \cite{kocmi2023large}, error levels \cite{lu2023error}, or fluency \cite{yang2023knowledge} using LLMs; second, leveraging the generative probabilities from LLMs, which involves using various prompts and examples to obtain multiple generative probabilities for the translated sentences of source texts, thereby calculating the mean and variance to gain a more accurate measure of uncertainty for translation quality \cite{huang2023towards}; third, generating pseudo data based on the knowledge within LLMs to then transfer to QE models \cite{xu2023instructscore,HUANG2024102022}; fourth, employing LLMs as pre-trained foundation models to enhance QE systems \cite{xu2023instructscore,gladkoff2023predicting}; and fifth, adopting retrieval-based approaches to infuse translation knowledge into LLMs \cite{huang2023towards,HUANG2024102022}. Although the performance of LLM-based QE methods has not yet surpassed that of QE methods incorporating pre-trained LMs, it is anticipated that with ongoing research, LLM-based approaches have the potential to reach state-of-the-art (SOTA) performance levels.

Indeed, despite the significant advancements made in QE, there remain several challenges that urgently need to be addressed, including data scarcity, insufficient interpretability, the rarity of word-level and document-level QE methods, the high computational resource requirements of pre-trained LMs and LLMs, and the lack of standardized evaluation benchmarks. To improve the accuracy, interpretability, and sustainability of QE, these challenges must be tackled one by one.

In this paper, our aim is to provide a clear and concise overview of MTQE for practitioners engaged in QE research and scholars interested in entering this field. In contrast to shared task overviews, our work not only synthesizes the WMT QE shared tasks from the past four years but also broadens the scope of content. Specifically, this paper reviews datasets, annotation methods, shared tasks, and all the seminal classic methods within the QE domain, with a particular emphasis on the currently highly-regarded QE approaches based on LLMs. Moreover, we explore the specific impact of LLMs on QE, a topic not yet covered in other survey reviews. Ultimately, we engage in an in-depth discussion of the current challenges faced by QE and the future research directions.

\subsection{Outline}

In Section \ref{dataset,annotations,tasks}, we discuss commonly used datasets for QE and categorize annotation methods into Human Translation Error Rate (HTER), Direct Assessment (DA), and Multi-dimensional Quality Metrics (MQM) based on their application scenarios. We also classify QE shared tasks into word-level, sentence-level, document-level and explainable QE. However, QE tasks are still evolving and there is a need for more reasonable objectives and data annotation principles.

In Section \ref{methods}, as shown in Fig.~\ref{fig:all}, we review different methods in the QE field and classify them into handcrafted features-based methods, deep learning-based methods, and LLM-based methods. Among deep learning-based methods, we further categorize them into classic deep learning methods and those incorporating pre-trained LMs. In Section \ref{finding}, we list five major challenges that currently exist in the QE field. Finally, in Section \ref{conclusion}, we provide our conclusions.

% \begin{figure*}
%     \centering
%     \includegraphics[width=\textwidth]{figures/all_figures_new.png}
%     \caption{All Methods Mentioned in This Paper. Due to Space Limitations, This Paper Only Provides Detailed Descriptions of the Method Highlighted in Bold.}
%     \label{fig:all}
% \end{figure*}

\subsection{Contribution}

Our contributions can be summarized as follows:

\begin{itemize}
    \item 
    We provide a clear and concise overview for practitioners engaged in QE and scholars interested in entering this field of study, covering the research development in QE, a significant and innovative area within NLP. This includes datasets, annotation methods, shared tasks, and nearly all the key methods within the QE domain, with a special emphasis on the currently popular QE approaches based on LLMs, a topic not yet covered in other survey reviews.

    \item 
    % We categorize the methods that have emerged throughout the development of the QE field into three main classes: methods based on handcrafted features, methods grounded in deep learning, and methods leveraging LLMs. 
    We classify the methods that have emerged throughout the development of the QE field into three main categories: those that employ handcrafted features, those grounded in deep learning, and those leveraging LLMs. We have conducted an in-depth exploration of nearly all the representative methods within the QE domain, placing particular emphasis on elucidating the intrinsic connections among them. Our goal is to provide a thorough and professional understanding of the current state of QE methodologies.

    \item 
    Compared to overviews of shared tasks, we have synthesized the QE shared tasks from WMT over the past four years, and included additional content. Furthermore, we delve into a discussion of the five challenges faced by QE, as well as prospective research directions for future.
    
\end{itemize}

\section{DATA, ANNOTATIONS METHODS, AND SHARED TASKS FOR QUALITY ESTIMATION} \label{dataset,annotations,tasks}

This section offers a comprehensive overview of QE, covering datasets, annotation methods, and shared tasks. It reviews datasets used in QE studies, explores annotation methods, and introduces shared tasks at word-level, sentence-level, document-level, and explainable QE. These aspects provide valuable resources and evaluation methods for researchers.

\subsection{Datasets}

The MLQE-PE dataset \cite{fomicheva2020mlqepe} is a significant milestone in QE and 
Automatic Post-Editing (APE) research, providing annotations in a multilingual environment. The dataset is constructed using sentences from Wikipedia and Reddit articles. Parallel corpora are generated for 11 different language pairs (LPs), including 7 conventional resource LPs (English-German -- En-De, English-Chinese -- En-Zh, Russian-English -- Ru-En, Romanian-English -- Ro-En, Estonian-English -- Et-En, Nepali-English -- Ne-En, and Sinhala-English -- Si-En) with 10K sentences each, divided into training, development, and two test sets (test20 and test21). Additionally, the dataset includes 4 zero-shot LPs (Pashto-English -- Ps-En, Khmer-English -- Km-En, English-Japanese -- En-Ja, and English-Czech -- En-Cs) with 2K sentences each, evenly split into two test sets too.

% The WMT2022 QE dataset, provided by WMT2022\footnote{https://wmt-qe-task.github.io/wmt-qe-2022/\label{wmt2022qe}}, includes annotations with HTER, DA, and MQM. It consists of 7 conventional resource LPs from the MLQE-PE \citep{fomicheva2020mlqepe}, where the training set includes MLQE-PE's training set, development set, and test20 portion. The development set uses the test21 portion of MLQE-PE. There are also 4 zero-shot LPs from MLQE-PE, where the development set and test set utilize the test20 and test21 portions, respectively. The dataset also introduces the En-Mr LP and includes MQM for En-De, En-Ru, and Zh-En LPs.

The WMT2023 QE dataset, provided by the organizers of WMT2023, encompasses DA and post-editing (PE) data, as well as data based on MQM. It is noteworthy that the data for English-Hindi -- En-Hi, English-Gujarati -- En-Gu, English-Tamil -- En-Ta, English-Telugu -- En-Te, English-Persian -- En-Fa, and Hebrew-English -- He-En LPs in the WMT2023 QE dataset are newly released in 2023.

The DA \& PE data includes all LPs from the MLQE-PE dataset and has added new LPs such as English-Yoruba -- En-Yo, English-Marathi -- En-Mr, En-Hi, En-Ta, En-Te, En-Gu, and En-Fa. Within this dataset, there are 14 LPs provided with PE information and 17 LPs with DA annotations. The training set comprises all LPs from the MLQE-PE dataset, with approximately 10,000 samples per LP; about 7,000 samples for En-Hi, En-Gu, En-Ta, and En-Te respectively; and approximately 27,000 samples for En-Mr. The test set includes LPs such as En-Mr, En-Hi, En-Gu, En-Ta, En-Te, and En-Fa, each with more than 1,000 samples.

The MQM data section covers four LPs: En-De, En-Ru, Zh-En, and He-En. The training set includes En-De, En-Ru, and Zh-En LPs with 30,425, 17,144, and 36,851 samples, respectively. The test set encompasses En-De, Zh-En, and He-En pairs, each with over 1,000 samples. 

These datasets provide an extremely important resource for research in the field of QE. They offer researchers corpora with rich texts and detailed annotations, which have propelled the progress of QE research. Specific related information can be found in Table \ref{tab:dataset}. It should be noted that, for the sake of brevity in presentation, the WMT2023 QE dataset omits all LPs from the MLQE-PE dataset.

\begin{table*}[h]
\caption{Datasets for QE. The two datasets listed are currently the most commonly used in the field of QE.}
\begin{tabularx}{\textwidth}{Xc@{\hskip 0.05in}c@{\hskip 0.05in}c@{\hskip 0.05in}c@{\hskip 0.05in}c@{\hskip 0.05in}c@{\hskip 0.05in}c@{\hskip 0.05in}c@{\hskip 0.05in}c@{\hskip 0.05in}c@{\hskip 0.05in}c@{\hskip 0.05in}c}
% \begin{tabularx}{\textwidth}{Xc@{\hskip 0.05in}c@{\hskip 0.05in}c@{\hskip 0.05in}c@{\hskip 0.05in}c@{\hskip 0.05in}c@{\hskip 0.05in}c@{\hskip 0.05in}c@{\hskip 0.05in}c@{\hskip 0.05in}c@{\hskip 0.05in}c@{\hskip 0.05in}c@{\hskip 0.05in}c}

\toprule
\multirow{2}{*}{\textbf{Dataset}} & \multirow{2}{*}{\textbf{LPs}} & \multicolumn{3}{c}{\textbf{Sentences}} & \multicolumn{3}{c}{\textbf{Tokens}} & \multicolumn{3}{c}{\textbf{Annotations}} 
 & \multirow{2}{*}{\textbf{Data Source}}  & \multirow{2}{*}{\textbf{Release Date}}\\ 
% \cmidrule(l){3-11}
\cmidrule(r){3-5} \cmidrule(lr){6-8} \cmidrule(){9-11}
 &  & \textbf{Train} & \textbf{Dev} & \textbf{Test} & \textbf{Train} & \textbf{Dev} & \textbf{Test}  &  \textbf{DA} & \textbf{PE} & \textbf{MQM} \\ 
 
 \hline
 \noalign{\smallskip}
\multirow{11}{*}{\textbf{MLQE-PE}}   \hspace{4pt} &
En-De & 7,000 & 1,000 & 1,000/1,000 & 114,980 & 16,519 & 16,371/16,545 & \checkmark & \checkmark & & Wikipedia & 2021/22  \\

 & En-Zh & 7,000 & 1,000 & 1,000/1,000 & 115,585 & 16,307 & 16,765/16,637 & \checkmark & \checkmark & & Wikipedia & 2021/22 \\
 & Ru-En & 7,000 & 1,000 & 1,000/1,000 & 82,229 & 11,992 & 11,760/11,650 & \checkmark & \checkmark & & Reddit & 2021/22  \\
 & Ro-En & 7,000 & 1,000 & 1,000/1,000 & 120,198 & 17,268 & 17,001/17,359 & \checkmark & \checkmark & & Wikipedia & 2021/22  \\
 & Et-En & 7,000 & 1,000 & 1,000/1,000 & 98,080 & 14,423 & 14,358/14,044 & \checkmark & \checkmark & & Wikipedia & 2021/22  \\
 & Ne-En & 7,000 & 1,000 & 1,000/1,000 & 104,934 & 15,144 & 14,770/15,017 & \checkmark & \checkmark & & Wikipedia & 2021/22  \\
 & Si-En & 7,000 & 1,000 & 1,000/1,000 & 109,515 & 15,708 & 15,821/15,709 & \checkmark & \checkmark & & Wikipedia & 2021/22  \\
 & Ps-En & -     & 1,000 & 1,000        & -       & 27,045 & 27,414       & \checkmark & \checkmark & & Wikipedia & 2021/22 \\ 
 & Km-En & -     & 1,000 & 1,000        & -       & 21,981 & 22,048       & \checkmark & \checkmark & & Wikipedia & 2021/22 \\ 
 & En-Ja & -     & 1,000 & 1,000        & -       & 20,626 & 20,646       & \checkmark & \checkmark & & Wikipedia & 2021/22 \\
 & En-Cs & -     & 1,000 & 1,000        & -       & 20,394 & 20,244       & \checkmark & \checkmark & & Wikipedia & 2021/22 \\
\hline
\noalign{\smallskip}
\multirow{11}{*}{\textbf{WMT2023 QE}}  \hspace{4pt} 
& En-Mr & 27,000 & 1,000 & 1,086 & 717,581 & 26,253 & 27,951 & \checkmark & \checkmark & & multi-domain/multi-corpus & 2023 \\
 & En-Hi & 7,000 & 1,000 & 1,074 & 181,336 & 25,943 & 28,032 & \checkmark & & & multi-domain/multi-corpus & 2023 \\
 & En-Gu & 7,000 & 1,000 & 1,075 & 153,685 & 21,238 & 23,084 & \checkmark & & & multi-domain/multi-corpus & 2023 \\
 & En-Ta & 7,000 & 1,000 & 1,067 & 150,670 & 21,655 & 20,342 & \checkmark & & & multi-domain/multi-corpus & 2023 \\
 & En-Te & 7,000 & 1,028 & 1,000 & 147,492 & 20,686 & 22,640 & \checkmark & & & multi-domain/multi-corpus & 2023 \\
 & En-Fa & - & - & 1,000 & - & - & 26,807 & & \checkmark & & news (multi-domain) & 2023 \\
 & En-De & 30,425 & - & 1,897 & 877,066 & - & 37,996 &  &  & \checkmark & multi-domain & 2021/23 \\
 & En-Ru & 17,144 & - & - & 395,045 & - & - &  &  & \checkmark & multi-domain & 2021/22 \\
 & Zh-En & 36,851 & - & 1,675 & 1,654,454 & - & 39,770 &  &  & \checkmark & multi-domain & 2021/23 \\
 & He-En & - & - & 1,182 & - & - & 35,592 &  &  & \checkmark & multi-domain & 2023 \\
\bottomrule
% \end{tabular}

\label{tab:dataset}
% \end{center}
\end{tabularx}
\end{table*}

\subsection{Annotation Methods}  \label{annotation}

This section discusses annotation methods in QE, which serve QE systems by providing labeled data. Three main methods will be introduced: HTER, DA, and MQM, each with its unique advantages and limitations, and suitable for different application scenarios. Finally, this section provides a discussion on the difficulty level associated with each of the three annotation methods.

\subsubsection{Human Translation Error Rate (HTER)}

HTER is a common method used for annotating translated sentences based on the volume of PE effort required. It builds upon word-level QE and is calculated from the results thereof. The goal of the reference translation is to make the least possible modifications while maintaining the original meaning and grammatical correctness. HTER scores a translated sentence by calculating the proportion of 
the number of edits (insertions, deletions, and replacements) made during the PE process to the number of words in the post-edition, with the formula provided in \eqref{eq1}. In previous research, HTER has been analyzed as a substitute for human assessment, with some studies recommending its use as a gold standard for evaluation. However, opinions vary within the academic community regarding the adequacy of HTER as a substitute.

\begin{equation}
HTER = \frac{\# \text{ of edits}}{\#\text{ of words in the post-edition}}\label{eq1}.
\end{equation}

\subsubsection{Direct Assessment (DA)} \label{DA}

DA is a widely utilized manual evaluation method that provides a subjective quality assessment, taking into account the overall effect of translation outputs and serving as an alternative to HTER. During the DA evaluation process, annotators rate the quality of translations directly within a range of 0-100. When utilizing multiple DA scores as targets for QE tasks, these scores are typically normalized first, and the normalized averages are then used to represent the quality score of the MT output. DA scores are prone to inconsistency due to the influence of annotators' individual preferences. However, solutions have been proposed to enhance the consistency among annotators \cite{guzman-etal-2019-flores}. As a result, DA has established itself as a reliable manual evaluation method and is extensively employed in QE tasks. Some advocate human assessments using DA, while others \cite{fomicheva2020mlqepe} believe that DA and HTER offer distinct perspectives on MT quality. Both viewpoints are considered valid.

\subsubsection{Multi-dimensional Quality Metrics (MQM)}

MQM \cite{lommel2014multidimensional} is an innovative and more objective annotation method that combines multiple evaluation indicators. It divides MT errors into 7 dimensions: Terminology, Accuracy, Linguistic conventions, Style, Locale conventions, Audience appropriateness, and Design and markup. Each dimension is subdivided into different error types, allowing for a more fine-grained QE of MT. Each dimension corresponds to four severity levels: no errors, minor errors, major errors, and critical errors, with different penalty scores for each level. Annotators can adjust parameters based on specific needs and integrate MQM into specific scenarios. Points are deducted based on the type of errors triggered by the MT, and the MT scores are calculated by subtracting the penalty scores from the full scores. MQM provides a more comprehensive and objective evaluation of MT quality compared to HTER and DA. It offers flexibility and personalized assessments, but requires annotators with domain knowledge and careful parameter settings. The formula is shown as in \eqref{eq2}, where $n_{\text{minor}}$, $n_{\text{major}}$, $n_{\text{critical}}$, and $n$ correspond to the counts of minor errors, major errors, critical errors, and the total number of words, respectively.

\begin{equation}
MQM = 1 \textendash \frac{n_{\text{minor}} \textendash 5n_{\text{major}} \textendash 10n_{\text{critical}}}{n}\label{eq2}.
\end{equation}

\subsubsection{Discussion of Difficulty}

HTER places high demands on the annotators' language skills and editing expertise, requiring a deep understanding of the linguistic characteristics of both the source and target languages. Its results may be influenced by the annotators' editing styles, leading to inconsistencies among different annotators. DA necessitates direct scoring by annotators, who need training to ensure consistent application of the scoring standards. MQM demands high levels of professional knowledge from annotators, who must have a thorough understanding of error types and perform precise annotations. 
% HTER and MQM emphasize fine-grained control over translation quality but can struggle with annotation inconsistency and efficiency. The simpler, faster DA approach has wider adoption yet faces subjectivity and variability between scorers, necessitating strict oversight. 
Overall, no one paradigm perfectly resolves the intrinsic trade-offs in MTQE. Ongoing research toward standardized best practices aims to combine these metrics' respective strengths.
% Overall, HTER and MQM emphasize fine-grained control over MT quality but may encounter issues with low annotator consistency and poor annotation efficiency. DA, with its simple and rapid operation, has been widely adopted, but its subjectivity and variability in annotator scoring require strict quality control.

\subsection{Shared Tasks}  \label{tasks}

QE shared tasks aim to advance the SOTA in QE by providing standardized datasets and evaluation metrics. These tasks cover QE at different levels and have diverse objectives. The popular QE shared tasks can be categorized into word-level, sentence-level, document-level, and explainable QE, each with its own objectives, evaluation metrics, and rationale.

\subsubsection{Word-level QE Shared Tasks}

The objective of word-level QE is to use words as the basic unit of assessment, automatically identifying the correctness of each word's position in a translated sentence, as well as detecting any mistranslation and omission phenomena, with reference to the source sentence. The input for this task includes the source text and the MT text, while the output is a series of labeled tag sequences (including source tags, MT tags, and gap tags). Each tag corresponds to each word or gap in the translated sentence, indicating whether there is an error at that location.

After summarizing the word-level QE shared tasks of the WMT in the past four years, we have categorized them into three types: classification, regression, and fine-grained error span detection. Classification tasks involve categorizing both the source and the target, with a further distinction between word classification and gap classification; translations that are correct are marked as OK, while those with errors are marked as BAD. Regression tasks employ semi-supervised or unsupervised models to score words based on sentence-level scores, with a threshold set to label words above it as OK and those below it as BAD. 
Fine-grained error span detection is a new task introduced in WMT2023 QE\footnote{https://wmt-qe-task.github.io/subtasks/task2/}, which categorizes translated words into no errors, minor errors, and major errors, and predicts error spans by linking indices of words within the same category through post-processing.

The primary evaluation metric for word-level QE is the Matthews correlation coefficient (MCC), accompanied by the F1-score as a secondary metric. MCC is particularly suited for binary classification models and datasets with an uneven distribution. It is used to measure the correlation between incorrectly translated words and manual annotations.

\subsubsection{Sentence-level QE Shared Tasks}

Sentence-level QE aims to predict the quality score for each LP, indicative of the translation quality, akin to a regression task in ML. It employs annotation methods such as HTER, DA, and MQM to assess the quality of translations. The sentence-level QE tasks at WMT2021\footnote{https://www.statmt.org/wmt21/quality-estimation-task.html}, WMT2022, and WMT2023\footnote{https://wmt-qe-task.github.io/subtasks/task1/} have employed HTER, DA, and MQM annotations.

The primary evaluation metric for sentence-level QE is the Spearman's rank correlation coefficient (Spearman's $\rho$), while the Pearson's and Kendall's correlation coefficients are used as auxiliary evaluation metrics. The Spearman's $\rho$ does not rely on assumptions of normality and equal variance in translation quality scores and is less affected by outliers. Thus, it provides a better reflection of the correlation between the translation quality predicted by the MT model and manual annotation.

\subsubsection{Document-level QE Shared Tasks}

Compared to finer-grained word-level and sentence-level QE, document-level QE is much more complex and requires a significant amount of data resources. The core objective of document-level QE is to perform QE on translation documents, where ``document" often refers to a text containing at least 3 sentences, rather than just a single document. Traditional MT tasks typically treat a single sentence as the basic unit of input and translation, overlooking the interdependence between sentences within a document. This approach may result in a lack of semantic coherence throughout the entire document. Since its development in 2016, document-level QE tasks have primarily focused on two types of prediction targets. One type involves calculating quality scores using two-step PE methods, while the other involves predicting MQM scores computed by MQM, as well as word-level and sentence-level error types.

Predicting both two-step PE scores and MQM scores use Pearson's correlation coefficient as the main evaluation metric, along with Mean Absolute Error (MAE) and Root Mean Square Error (RMSE) as supplementary metrics. On the other hand, predicting word-level error types uses F1-score as the evaluation metric.

\subsubsection{Explainable QE Shared Tasks}

In QE, interpretability is important for enhancing user trust and facilitating error analysis. Unlike sentence-level QE, which focuses on overall quality scores, explainable QE is concerned mainly with errors in the translation. This paper categorizes explainable QE into two scenarios. The first scenario aims to predict sentence-level binary scores to signal whether the translation contains critical errors. These errors, mainly caused by mistranslations, hallucinations, and deletions of content from the source sentence, could potentially lead to misinformation in areas like health, safety, law, reputation, and religion. Based on the scores, users can determine whether critical errors have occurred in the translation. The second scenario provides sentence-level quality scores that signal the presence of translation errors in the sentence but do not identify which specific words are mistranslated. These scores help users understand why a sentence may be deemed low quality.

In explainable QE, Recall at Top K is the primary evaluation metric, which measures the model's ability to detect and rank mistranslated words within the top K predictions made by the MT model. Area Under the Curve (AUC) and Average Precision are used as auxiliary evaluation metrics. \\

In conclusion, QE shared tasks are designed with different objectives in mind, focusing on defining quality indicators across various aspects. Each task is equipped with unique evaluation metrics to measure model performance. Word-level QE is akin to classification tasks, where words are labeled as OK or BAD. Sentence-level QE is similar to regression tasks, aiming to predict the quality scores for translated sentences. Document-level QE is more complex, responsible for scoring entire translated documents or text blocks containing multiple sentences. Explainable QE, on the other hand, is mainly concerned with errors in the translation, rather than the quality score of the translation. It not only identifies specific types of errors but also points out the words where translation errors exist based on the scores given to sentences, although it does not specify exactly which word is erroneous.

% The IEEEtran class file is used to format your paper and style the text. All margins, 
% column widths, line spaces, and text fonts are prescribed; please do not 
% alter them. You may note peculiarities. For example, the head margin
% measures proportionately more than is customary. This measurement 
% and others are deliberate, using specifications that anticipate your paper 
% as one part of the entire proceedings, and not as an independent document. 
% Please do not revise any of the current designations.
% IEEEtran 类文件用于格式化您的论文和样式化文本。所有的边距、栏宽、行间距和文本字体都已预先规定；请不要更改它们。您可能会注意到一些特殊之处。例如，头部边距比通常的比例要大。这种测量以及其他测量是故意的，使用的规格是预期您的论文作为整个会议记录的一部分，而不是作为一个独立的文档。请不要修改任何当前的设计。

\section{Methods of Quality Estimation}  \label{methods}
% Before you begin to format your paper, first write and save the content as a 
% separate text file. Complete all content and organizational editing before 
% formatting. Please note sections \ref{AA}--\ref{SCM} below for more information on 
% proofreading, spelling and grammar.

% Keep your text and graphic files separate until after the text has been 
% formatted and styled. Do not number text heads---{\LaTeX} will do that 
% for you.
% 在文本被格式化和样式化之后，再将您的文本和图形文件合并。

This section reviews relevant research work within the three main categories of methods that have emerged throughout the evolution of QE. It discusses the advantages and limitations of the respective methods within each category and provides a brief comparison between different approaches. We have categorized these methods just to facilitate the description of the development trend of QE, and cannot guarantee there is no overlap between the methods in different categories.

\subsection{Quality Estimation Based on Handcrafted Features} 
% Define abbreviations and acronyms the first time they are used in the text, 
% even after they have been defined in the abstract. Abbreviations such as 
% IEEE, SI, MKS, CGS, ac, dc, and rms do not have to be defined. Do not use 
% abbreviations in the title or heads unless they are unavoidable.
% 在文本中首次使用缩写和首字母缩略词时定义它们，即使它们已在摘要中定义过。像IEEE、SI、MKS、CGS、ac、dc和rms这样的缩写不需要定义。除非不可避免，否则不要在标题或标题中使用缩写。

Before 2009, QE research primarily focused on predicting quality labels for the output of Statistical Machine Translation (SMT) using handcrafted features \cite{blatz2004confidence,ueffing2007word}. Subsequently, the focus of QE research shifted towards predicting human-annotated quality scores. For instance, the QuEst \cite{specia-etal-2013-quest} framework 
% (see Figure 2 in Appendix A.1)
utilized a feature extraction module to extract quality labels from the source and translated text. These features 
% (see Table 7 in Appendix A.2)
were then applied to ML algorithms to construct QE systems. de Souza et al. \cite{de2014fbk} used a supervised tree-based ensemble learning method to predict PE effort and time under various features and BLSTM-RNNs to predict word-level labels.

QuEst++ \cite{specia2015multi} is an improved and expanded version of QuEst, with added feature extraction modules designed for word-level and document-level QE. It integrates predictions at three different levels into a single workflow, facilitating interactions between word-level, sentence-level, and document-level QE. Additionally, QuEst++ incorporates sequence labeling learning algorithms for word-level QE. This tool can be conveniently extended with new features to meet the requirements of different text levels, offering high flexibility. 
% For other notable methods, please refer to the content in Table \ref{tab:human-crafted methods}.

\subsection{Quality Estimation Based on Deep Learning}

Since the 2010s, deep learning technologies have been widely applied in the field of NLP, and from around 2015, they began to be integrated into QE methodologies. These approaches can be categorized into those based on classic deep learning techniques and those incorporating pre-trained LMs.

\subsubsection{Classic Deep Learning Methods}

With the advancement of QE, the advent of word embeddings \cite{mikolov2013efficient,pennington2014glove} and neural machine translation (NMT) \cite{bahdanau2014neural,sutskever2014sequence} technologies has led some researchers to apply neural networks to QE tasks. The progression from initially utilizing neural networks for feature extraction to the emergence of fully neural network-based QE systems has greatly enhanced the performance of QE systems.

Besides utilizing the handcrafted features from QuEst, SHAH et al. \cite{shah2015shef,shah2015investigating} also employed additional word-level QE features extracted from Word2Vec \cite{mikolov2013efficient} embeddings and the similarity in the embedding space between source and target language words. They combined language model probabilities generated from trained continuous space models with these handcrafted features for sentence-level QE. Furthermore, Scarton et al. proposed word embedding features \cite{scarton2016word}, discourse features, and features extracted from pseudo-reference translations \cite{scarton2015searching} for document-level QE. Inspired by their work, Chen et al. \cite{chen2017improving} proposed the use of sentence embedding features and cross-entropy features to enhance the correlation of QE with human evaluations and investigated several factors affecting the performance of QE systems.

Subsequently, some researchers explored the use of neural networks solely for feature extraction and QE. QUETCH \cite{kreutzer-etal-2015-quality} is an early example of this approach, utilizing pretrained word representations and a Deep Neural Network (DNN) architecture. The QUETCH
% as shown in Figure 3 of Appendix A.1,
comprises an input layer, lookup tables, a multilayer perceptron (MLP), and an output layer. It feeds bilingual context representations through a fixed-size word window into the MLP, ultimately completing the word-level QE task via the output layer. However, its effectiveness did not match that of QUETCH+, which integrated additional baseline features. Expanding on QUETCH, Martins et al. \cite{martins-etal-2016-unbabels} introduced a 200-unit bi-directional Gated Recurrent Unit (BiGRU) network and stacked feed-forward neural networks, subsequently incorporating POS tags for both source and target words to achieve the best performance of that time. Similar to QUETCH, Patel et al. \cite{patel2016translation} made a switch from DNN to RNN, utilizing LSTM and GRU for extracting representations of bilingual sequences, and introduced sub-labels to address the challenge of label imbalance.

While the QUETCH \cite{kreutzer-etal-2015-quality} approach relies entirely on neural networks for feature extraction, it requires bilingual alignment information, which is typically obtained via statistical methods and is prone to significant errors. With the advancement of deep learning technologies, the trend in QE research has been gradually shifting towards completely neural network-based methods.

In 2016, Kim et al. \cite{kim-lee-2016-recurrent,kim2016recurrent} made the first attempt to use an NMT model for QE, proposing the inaugural purely neural approach for sentence-level, word-level, and phrase-level QE that does not require manually extracted features. In 2017, Kim et al. \cite{kim2017predictor} presented a more in-depth study and named it the Predictor-Estimator (PredEst) model, which is a method to address the issues of expensive QE annotations and limited annotated QE data. It consists of two components: the predictor and the estimator. The predictor is a neural word prediction model trained using parallel corpora. It masks the target word, feeds the source language and the corrupted target language into bidirectional RNN (Bi-RNN), and predicts the probability distribution of the masked word. On the other hand, the estimator is a neural QE model trained on QE data, extracting QE feature vectors (QEFVs) and training them on a feedforward network. QEFVs are processed by FNN, RNN, or Bi-RNN to obtain hidden representations, which are then used to predict quality labels for sentence, phrase, or word-level tasks. Later, to train the model effectively, Kim et al. \cite{kim-etal-2017-predictor} introduced stack propagation and multi-level task algorithms to improve the original method. In 2018, Ive et al. \cite{ive2018deepquest} proposed the deepQuest framework for sentence-level and document-level QE, marking the first purely neural approach for document-level QE, which for the first time attempted to experiment with the outputs of both SMT and NMT. Upon testing, the framework proved to be faster and more cost-effective, and greatly improved the performance of the document-level QE framework.

Martins et al. \cite{martins2017pushing,martins2017unbabel} introduced a STACKEDQE system that stacked both linear and neural systems in the WMT17 word-level QE task, followed by combining APE with word-level QE to create an APEQE system. Ultimately, they merged these two systems to form the FULLSTACKEDQE system tailored for word-level QE and extended the FULLSTACKEDQE to sentence-level QE. These systems all achieved commendable results. Building on the approach of Martins et al. \cite{martins2017pushing,martins2017unbabel}, Hokamp et al. \cite{hokamp2017ensembling} incorporated features that had been demonstrated to be effective for word-level QE into the input of an NMT system, thus proposing the APE-QE model. This unified APE with word-level QE models and achieved the best performance of the time in both APE and QE tasks.

As the Transformer \cite{DBLP:journals/corr/VaswaniSPUJGKP17} model has garnered significant success in the field of MT, Fan, Wang, et al. \cite{DBLP:journals/corr/abs-1807-09433} developed a Bilingual Expert model based on a bidirectional Transformer and a PredEst architecture that includes a word prediction module and a QE module. The word prediction module leverages prior knowledge obtained from pretraining on a large parallel corpus and the joint latent representations between the source language and the translation for token prediction, extracting a set of features. Then, they introduce mismatch features that measure the discrepancy between the prior knowledge obtained from well-trained Bilingual Expert and the targets in the QE dataset to train the QE module, which uses a bidirectional LSTM model, achieving SOTA performance at that time. Wang et al. \cite{wang-etal-2020-hw-tscs} employed a pre-trained Transformer as the predictor and integrated Bottleneck Adapter Layers (BAL) for efficient transfer learning, with specific classifier and regressor as the estimator. They also conducted joint training for word and sentence-level tasks using a unified model and proposed a pseudo PE assisted QE method. This demonstrated the effectiveness of using pre-trained NMT models for transfer learning in QE tasks.

However, Cui et al. \cite{cui2021directqe} argued that the gap between data quality and training objectives in the PredEst framework hindered its ability to benefit from parallel corpora. Consequently, they proposed a framework called DirectQE, which includes a generator for creating pseudo QE data and a detector pre-trained with these pseudo data. This framework allows for pre-training using large parallel corpora and fine-tuning on real QE data, thereby addressing the issues inherent in the PredEst framework.

\subsubsection{Incorporating Pre-trained Language Models Methods}

With the emergence and development of pre-trained LMs such as ELMo \cite{peters2018deep}, BERT \cite{devlin2018bert}, XLM \cite{lample2019cross}, and XLM-R \cite{conneau2019unsupervised}, some studies have begun to integrate pre-trained LMs into QE models. This integration has enabled better extraction of quality vectors from source texts and translated texts, thereby enhancing the performance of QE systems.

Kepler et al. \cite{kepler2019unbabel} expanded OpenKiwi \cite{kepler2019openkiwi} into a Transformer-based PredEst model, replacing the predictor with the pre-trained LMs BERT and XLM, and proposed an ensemble method using the POWELL technique to combine word-level and sentence-level predictions. Moreover, they suggested a simple technique for converting word labels into document-level predictions. Wu et al. \cite{wu2020tencent} extended the OpenKiwi by integrating PredEst models based on XLM and Transformer in their submission to WMT20. The former predictor generates both masked and nonmasked representations, while the latter produces only nonmasked representations. The estimator is trained using either LSTM or Transformer, employing top-K and multi-head attention strategies to enhance sentence feature representation. Ranasinghe et al. \cite{DBLP:journals/corr/abs-2011-01536} proposed TransQuest, a PredEst model aimed at reducing the dependence of sentence-level QE on large-scale parallel corpora. TransQuest does not use parallel data to pre-train the predictor but instead directly employs SOTA cross-lingual embedding models like XLM-R \cite{conneau2019unsupervised,DBLP:journals/corr/abs-1911-02116} to encode the source and target sentences. It consists of two neural networks: MonoTransQuest (MTransQuest) and SiameseTransQuest (STransQuest). MTransQuest uses a single XLM-R model to encode concatenated source and target sentences, whereas STransQuest adopts a Siamese architecture, using separate XLM-R models for the source and translation. Both models use mean squared error loss as the objective function and have shown improved results with specific pooling strategies. 
Zerva et al. \cite{zerva2021unbabel} used a pre-trained multilingual encoder combined with adapters, training multilingual models on the OpenKiwi \cite{kepler2019openkiwi} PredEst and found that adapter tuning can resist overfitting. Additionally, they demonstrated that integrating uncertainty information and using out-of-domain data for pre-training can improve QE system performance.

Zhou et al. \cite{zhou2019source} primarily investigated the application of pre-trained translation models in QE and compared the effectiveness of a bilingual expert, ELMo, and BERT on QE tasks. Yankovskaya et al. \cite{yankovskaya2019quality} contrasted two approaches: one using only BERT and LASER \cite{artetxe2019massively} embeddings as features, and the other additionally incorporating log probability features from MT systems. Their research demonstrated the importance of the log probabilities from MT systems. 
% Ding et al. \cite{ding-etal-2021-levenshtein} employed the Levenshtein Transformer to replace the Masked Language Model (MLM) in the PredEst framework, proposing the use of data synthesis, heuristic subword-level reference, and pre-trained translation models to transfer translation knowledge to word-level QE tasks, achieving promising results in an unconstrained setting.

In 2020, Rei et al. \cite{rei2020comet} introduced COMET, a neural framework for training multilingual and adaptable MT evaluation models, which is often used for reference-based evaluation to generate predictive estimates of human judgments such as HTER, DA, and MQM. The COMET framework supports two different architectures: an Estimator model and a Translation Ranking model, both consisting of a cross-lingual encoder and pooling layers, with the fundamental difference being the training objective. The Estimator model, which is the most commonly used, is trained to directly regress to a quality score, while the Translation Ranking model is trained to minimize the distance between a ``better" hypothesis and its corresponding reference translation and source language. 

In 2022, Rei et al. \cite{rei2022cometkiwi} combined the strengths of COMET and OpenKiwi \cite{kepler2019openkiwi} by connecting COMET with the PredEst architecture of OpenKiwi, and equipped it with a word-level sequence tagger and explanation extractor, forming COMETKIWI for QE. COMETKIWI pre-trained the model on metrics data with the learning objective proposed by UniTE model \cite{wan2022unite}, which incorporates reference translation into the training, serving as a form of data augmentation. Additionally, COMETKIWI proposed an interpretability method using attention and gradient information and further refined the influence of attention heads on predictions through a Head Mix module. COMETKIWI also demonstrated the effectiveness of few-shot learning, achieving significant improvements in model performance with only 500 samples.

\subsection{Quality Estimation Based on Large Language Models}

With the development of LLMs, more researchers are turning their attention to utilizing LLMs for QE. The current methods can be roughly divided into five types, which have contributed significantly to the advancement of QE.

\subsubsection{Direct prediction based on content generated by LLMs}
    Kocmi and Federmann \cite{kocmi2023large} proposed GEMBA, a GPT-based \cite{brown2020language} translation quality assessment metric that uses single-step prompting and can be applied to scenarios with reference translations as well as to QE. They evaluated 9 different GPT models and concluded that only GPT-3.5 and larger models are capable of performing QE. GEMBA focuses on zero-shot prompting, with the authors employing 4 different prompt templates to execute quality assessments for both reference-based and nonreference-based translation modes. GEMBA predicts scores directly based on content generated by LLMs, assessing each segment independently, then averaging the scores of all segments to obtain a final system-level score, achieving SOTA performance at the system level, but lacking in segment-level analysis. To enhance the performance of LLMs in quality assessment, Lu et al. \cite{lu2023error} introduced Error Analysis Prompting (EAPrompt), a new prompting method that combines Chain-of-Thought (CoT) \cite{wei2022chain} with EA \cite{lu2022toward}. Using ChatGPT, this approach predicts the degree and number of errors and provides a score based on the severity of those errors, creating MQM-like assessments. It achieved better results on GPT-3.5-turbo than GEMBA. 
    Yang et al. \cite{yang2023knowledge} introduced the Knowledge-Prompted Estimator (KPE), which is a CoT approach that integrates three single-step prompting techniques, using LLMs to predict fluency, word-level similarity, and sentence-level similarity \cite{yang2023teachersim}, resulting in better performance for segment-level QE. Furthermore, KPE also demonstrated its advantages in terms of interpretability.

\subsubsection{Based on the generative probabilities of LLMs}
    Huang et al. \cite{huang2023towards} proposed an unsupervised QE framework that can explore useful information in LLMs. This framework makes use of both the sequence-level probability and the uncertainty of LLMs as evidence for how well LLMs perform QE. Specifically, they designed translation style prompts to guide LLMs in generating sequence-level probability, while they also investigated the impact of LLM's uncertainty on sequence-level probability. This was done by designing various prompts and demonstrations, and feeding the same sample into the LLMs multiple times in different forms, which resulted in multiple distinct generation probabilities. Through their analysis, they found that prompts of translation style and successful demonstrations can enhance the performance of the model.

\subsubsection{Leveraging LLMs to generate pseudo data}
    Xu et al. \cite{xu2023instructscore} introduced INSTRUCTSCORE, a method that learns interpretable text generation metrics without the need for human-annotated scores. This method constructs MQM-like data using knowledge provided by GPT-4 \cite{openai2023gpt4} to train the LLaMA model \cite{touvron2023llama}. First, they use GPT-4 to build pseudo data with errors and explanations, which is then used for fine-tuning LLaMA model. After that, they sample and input real-world diagnostic output into GPT-4 for feedback,  and then fine-tune the LLaMA model further by choosing the explanation most consistent with human input. They repeat these latter two steps to optimize the model's output. Ultimately, they found that the model performs well across tasks, domains, dimensions, and even on unseen tasks. Huang et al. \cite{HUANG2024102022}, on the other hand, leveraged LLMs to corrupt reference sentences. They then generated a fluent sentence from the corrupted one and used this as a noisy negative view. This approach, since it does not require data annotation, has strong generalization capabilities.

\subsubsection{LLMs as the foundation for QE models}
    Gladkoff et al. \cite{gladkoff2023predicting} fine-tuned LLMs using the OpenAI API interface to assess whether translations require post-editing. Their results showed that increasing the size of LLMs does not significantly improve the performance of the evaluation task. Similarly, Xu et al. \cite{xu2023instructscore}, as mentioned earlier, also utilized pseudo data generated by GPT-4 \cite{openai2023gpt4} to train the LLaMA model \cite{touvron2023llama}.

\subsubsection{Retrieval-based methods}
    This is a supplementary augmentation strategy. As previously mentioned, Huang et al. \cite{huang2023towards,HUANG2024102022} utilized BM25 \cite{robertson2009probabilistic} to retrieve samples that are highly similar to the demonstrations and require evaluation. This served to enrich the demonstration set, thereby enhancing the knowledge of LLMs.

\section{Finding} \label{finding}

Based on our observations of these methods, we have identified the following findings regarding the current challenges and developments in QE:

\begin{itemize}
    \item Data Scarcity: There is a scarcity of manually annotated data, particularly for low-resource languages. Acquiring sufficient annotated data involves significant costs, which to a large extent hinders the progress of QE research.

    \item Insufficient Interpretability: Earlier QE methods lacked interpretability, making it difficult to identify specific types of errors and their locations. In contrast, LLMs possess a strong knowledge base and learning capabilities. Future research should focus more on leveraging LLMs for enhancing the interpretability of QE.

    \item Word-level and document-level QE methods are rare. Current QE approaches predominantly focus on sentence-level, with limited work targeting word-level and document-level QE, particularly with word-level methods being few in number and lacking in performance. However, word-level QE can extract more fine-grained information, and future research should pay more attention to word-level QE.

    \item Pre-trained LMs and LLMs require a lot of hardware resources. Many research teams are unable to independently pre-train LMs due to a lack of sufficient hardware resources, forcing them to rely on open-source pre-trained LMs, which impedes the development of QE.

    \item Lack of standardized evaluation metrics: Due to the subjectivity of the QE task and varying preferences for translation quality, the absence of uniform evaluation metrics makes it challenging to compare and integrate model performance.
    
\end{itemize}

\section{Conclusion}   \label{conclusion}

Over the past 20 years, significant progress has been made in QE. As an application that can evaluate the quality of translated texts in real time without the need for reference translations, QE has strong practicality and plays a significant role in advancing the development of MT. This paper provides a comprehensive introduction and analysis of QE, offering an extensive overview of datasets, annotation methods, shared tasks, and methodologies. Specifically, the paper presents the specific concepts and details of word-level, sentence-level, document-level, and explainable QE shared tasks. It categorizes the methods developed throughout the evolution of QE into those based on handcrafted features, those grounded in deep learning, and those leveraging LLMs, further subdividing the deep learning-based methods into classic deep learning and those incorporating pre-trained models. This paper provides a detailed account of the advantages and limitations of each type of method and offers a simple comparison of different approaches. Finally, the paper discusses the current challenges in the QE field and suggests future research directions.

% \section*{References}

% \bibliographystyle{IEEEtran}
% \bibliography{IEEEabrv,reference}
\bibliographystyle{IEEEtran}
\bibliography{IEEEabrv,reference}

% \bibitem{b1} G. Eason, B. Noble, and I. N. Sneddon, ``On certain integrals of Lipschitz-Hankel type involving products of Bessel functions,'' Phil. Trans. Roy. Soc. London, vol. A247, pp. 529--551, April 1955.
% \bibitem{b2} J. Clerk Maxwell, A Treatise on Electricity and Magnetism, 3rd ed., vol. 2. Oxford: Clarendon, 1892, pp.68--73.
% \bibitem{b3} I. S. Jacobs and C. P. Bean, ``Fine particles, thin films and exchange anisotropy,'' in Magnetism, vol. III, G. T. Rado and H. Suhl, Eds. New York: Academic, 1963, pp. 271--350.
% \bibitem{b4} K. Elissa, ``Title of paper if known,'' unpublished.
% \bibitem{b5} R. Nicole, ``Title of paper with only first word capitalized,'' J. Name Stand. Abbrev., in press.
% \bibitem{b6} Y. Yorozu, M. Hirano, K. Oka, and Y. Tagawa, ``Electron spectroscopy studies on magneto-optical media and plastic substrate interface,'' IEEE Transl. J. Magn. Japan, vol. 2, pp. 740--741, August 1987 [Digests 9th Annual Conf. Magnetics Japan, p. 301, 1982].
% \bibitem{b7} M. Young, The Technical Writer's Handbook. Mill Valley, CA: University Science, 1989.
% \end{thebibliography}
% \vspace{12pt}
% \color{red}
% IEEE conference templates contain guidance text for composing and formatting conference papers. Please ensure that all template text is removed from your conference paper prior to submission to the conference. Failure to remove the template text from your paper may result in your paper not being published.

\end{document}